\documentclass[a4paper, times, 10pt, twocolumn, twoside,top=3.5cm,bottom=3.5cm,left=2.5cm,right=2cm]{article}

\usepackage{ISARC}
\usepackage{graphicx,subfigure}
\usepackage{color}
\usepackage{CJK}
\usepackage{framed}
\usepackage{ulem}
\usepackage{float}
\usepackage{cite}

\begin{document}


\definecolor{shadecolor}{rgb}{1,1,0}
\linespread{0.5}
\title{Building an Integrated Mobile Robotic System for Real-Time Applications in Construction}

\author{Khashayar Asadi$^{a}$, Hariharan Ramshankar$^b$, Harish Pullagurla$^b$, Aishwarya Bhandare$^b$, \\ \textbf{\large{Suraj Shanbhag$^b$, Pooja Mehta$^b$, Spondon Kundu$^b$, Kevin Han$^c$, Edgar Lobaton$^d$, Tianfu Wu$^e$}}}
\affiliation{$^{a,c}$Department of Civil, Construction, and Environmental Engineering, North Carolina State University, US\\
$^{b,d,e}$Department of Electrical and Computer Engineering, North Carolina State University, US}
\email{kasadib@ncsu.edu}

\maketitle 
\thispagestyle{fancy} \pagestyle{fancy}

\begin {abstract}
One of the major challenges of a real-time autonomous robotic system for construction monitoring is to simultaneously localize, map, and navigate over the lifetime of the robot, with little or no human intervention. Past research on Simultaneous Localization and Mapping (SLAM) and context-awareness are two active research areas in the computer vision and robotics communities. The studies that integrate both in real-time into a single modular framework for construction monitoring still need further investigation. A monocular vision system and real-time scene understanding are computationally heavy and the major state-of-the-art algorithms are tested on high-end desktops and/or servers with a high CPU- and/or GPU- computing capabilities, which affect their mobility and deployment for real-world applications. To address these challenges and achieve automation, this paper proposes an integrated robotic computer vision system, which generates a real-world spatial map of the obstacles and traversable space present in the environment in near real-time. This is done by integrating contextual Awareness and visual SLAM into a ground robotics agent. This paper presents the hardware utilization and performance of the aforementioned system for three different outdoor environments, which represent the applicability of this pipeline to diverse outdoor scenes in near real-time. The entire system is also self-contained and does not require user input, which demonstrates the potential of this computer vision system for autonomous navigation.
\end{abstract}


\section{Introduction}
\label{sec:intro}
A mobile robot operating in the physical world must be aware of its environment. A large part of this awareness is about estimating spaces (i.e., of mapping) and the robot's location (i.e., localization) \cite{makarenko2002experiment}. In the absence of external localization aids, the robot must be able to build a map and, at the same time, localize itself in the same partially built imperfect map \cite{leonard1991simultaneous}. The robot must be "contextually aware" of its surroundings, meaning that the robot must be capable of sensing different objects and making situation-specific decisions based on them. This is achieved through object recognition via semantic segmentation, which enables the generation of a spatial map of the obstacles and the traversable space of the environment. This work can also boost autonomous robotic applications to achieve a higher degree of automation in construction monitoring and personalized safety, which have high rate of interest among researchers in this area \cite{han2017potential,asadi2018real,boroujeni2017perspective,Siebert2014uav,KROPP2018image,jeelani2016development,shakeri2015lean,mahmoodzadeh2015energy,keshavarzi2016viscoelastic}. 

\begin{figure}[H]
\centering
\includegraphics[width=0.45\textwidth]{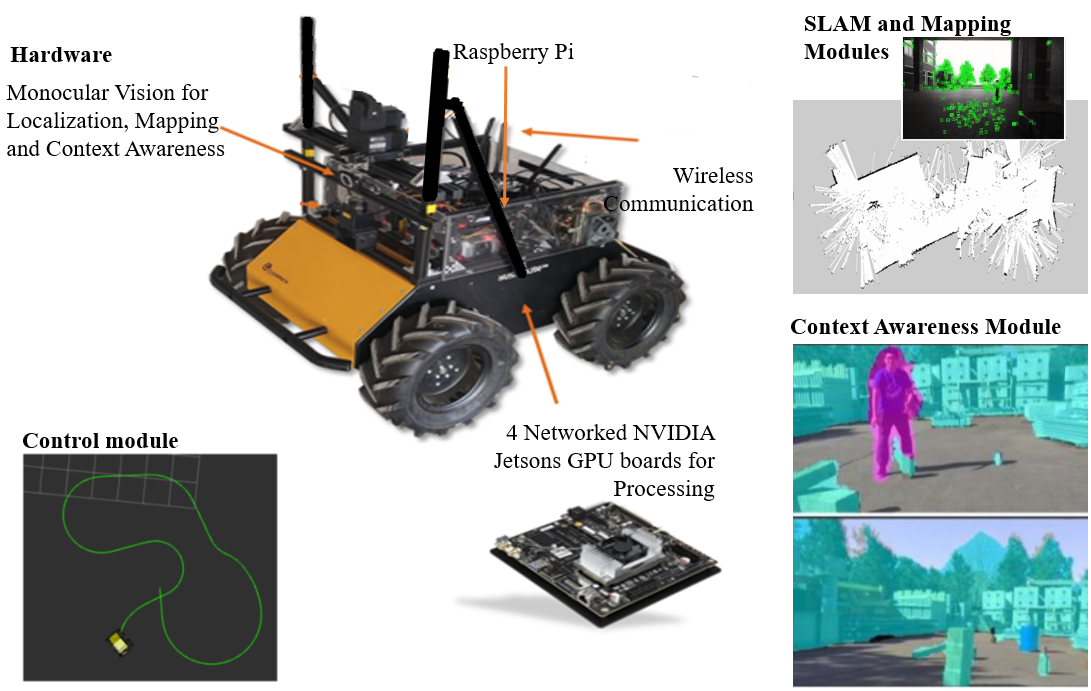}
\caption{Overview of system components-hardware, control, SLAM, context awareness and mapping}
\label{fig:overview}
\end{figure}

The recent advance in powerful and portable processing units have enabled analysis of complex data streams in real-time. These small processing units with high-computing capabilities are well-suited for environmental monitoring using a combination of cameras, microphones, and sensors for temperature, air-quality, and pressure \cite{betthauser2014wolfbot,wilson2016pheeno,gholizadeh2017fault}. Still, there are a few well-set platforms that combine the state-of-the-art hardware with accessible software and opensources. This paper proposes an integrated mobile robotics agent that is capable of processing localization, mapping, scene understanding, and control and planning. The proposed integrated system uses multiple NVIDIA Jetson TX1 boards \cite{NVIDIA}, each handling a specific task. The low-power consumption and integrated GPU make the Jetson TX1 an ideal candidate for running the aforementioned processes in real-time. As illustrated in Figure \ref{fig:overview}, the sub-tasks are control, simultaneous localization and tracking (SLAM), image segmentation (denoted as Context Awareness), mapping. For validation, three case studies that captures three different outdoor scenes are performed to evaluate the system's robustness, performance, its integration, and, therefore, its feasibility of the future development of an autonomous ground robot. 

This paper is organized as follows: Section 2 presents a literature review on SLAM and scene understanding which are the primary sub-tasks in this approach. Sections 3 describes the system overview and hardware modules of the proposed system. Section 4 describes the approaches taken to perform the above-mentioned pipeline. Section 5 presents the system evaluation and results. Section 6 ends this paper with the conclusions and future works. 

\section{Background}
\subsection{Monocular SLAM}
In a dynamic environment, human eyes can quickly and accurately provide visual information that our brain uses to map and understand the environment. A robot must also know with a high degree of certainty, where it is located in the environment. Only if this occurs, can it localize itself with respect to the map, which is essential for tasks such as navigation and motion planning. Likewise, using cameras, robots get visual information and they can generate maps of their environment. For building a good map, accurate localization is needed and for accurate localization, a good map is necessary. This chicken and egg problem is what SLAM aims to solve \cite{durrant2006simultaneous}. 

Initial approaches focused on 2D maps using a laser scanner or LIDAR \cite{fontanelli2007fast}. The advantage of such approaches was the speed of mapping and the lower computational cost. However, LIDAR-based approaches suffer when there is a lot of heat and reflective surfaces that affect the laser \cite{gao2016smartphone}. Also, 2D approaches are unable to capture the scale of the obstacles. With an increase in processing power and development of better algorithms, the use of cameras for SLAM became more feasible.

When it comes to monocular vision-based SLAM, ORB-SLAM \cite{mur2015orb}, Direct Sparse Odometry (DSO) \cite{engel2017direct} and LSD-SLAM \cite{engel2014lsd} are the widely used algorithms. ORB-SLAM is a feature-based method, while LSD-SLAM is a direct method based on color intensities in the image. Relying on feature extraction, feature matching, and visual odometry, maps are built, but the drawback is that the map is accurate only up to a scale. There are several methods available in the literature which can achieve this task in real-time \cite{mur2015orb,mur2016orb,engel2014lsd,forster2014svo,concha2015dpptam}. The approach discussed in \cite{mur2015orb} is the most appropriate for the current task due to its speed and ability to run in real-time on the TX1. The point cloud and odometry of ORB-SLAM is not directly usable as their units are not in real-world scale. Hence, one of the tasks is to transform the unscaled odometry to real-world units.  In this work, the SLAM Module provides the odometry and tracking state to the Context-Awareness and Control Modules.

\subsection{Scene Understanding}
The literature in object recognition is very rich, and yet growing. The goal of making the robot contextually aware can be achieved through recognition of the objects and taking decisions based on such insights. Either object classification or scene segmentation can be used for scene understanding purposes. Scene segmentation, while being more computationally intensive, provides more precise results, especially near the boundaries of objects.

Even though, convolutional neural networks (CNNs) had been utilized for a long time \cite{lecun1998gradient}, they seemed to be hard to use for bounding-box object classification up until 2014. This was when an image region proposal scheme was combined with CNN's as classifiers and was able to outperform other object detection frameworks \cite{girshick2014rich}. Later versions of R-CNN object detection were introduced to resolve some of the R-CNN's limitation, such as training pipeline complexity and slow test-time \cite{ren2015faster,girshick2015fast}. Fast R-CNN \cite{girshick2015fast} sped up the inference time by a factor of 25. In this method, computation of convolutional layers was shared between region proposals of an image. Faster R-CNN \cite{ren2015faster} inserted a region proposal network (RPN) after the last convolutional layer. By this change, the method required no external region proposal which improved computational speed up to 250x. 

High computational load is one of the main limitations for (CNN)-based frameworks for semantic segmentation. \cite{long2015fully} proposed a semantic segmentation framework using fully convolutional networks and utilizing existing classification networks, such as GoogLeNet \cite{szegedy2015going} and AlexNet \cite{krizhevsky2012imagenet}. This method transfers learning approaches via fine-tuning of pre-trained models. \cite{badrinarayanan2017segnet} is also based on a very large encoder-decoder model performing pixel-wise labeling which suffers from a tralarge number of computations.

MobileNets \cite{howard2017mobilenets} and Enet \cite{paszke2016enet} are computationally lighter convolutional networks. ENet is designed to run on embedded boards with a focus on distinguishing roads from the rest of the scene. These capabilities make it suitable for autonomous robot navigation in outdoor (and particularly construction) sites. 

\begin{figure*}[bt]
\centering
\includegraphics[width=\textwidth]{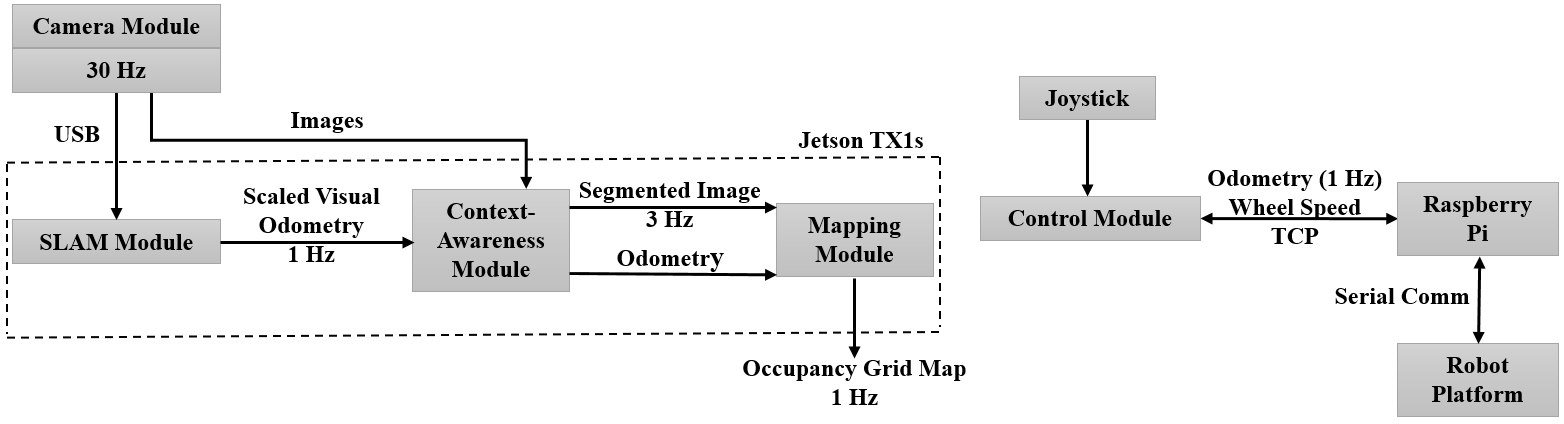}
\caption{Monocular vision-based general pipeline}
\label{fig:Flowchart}
\end{figure*}

\section{System Overview}
The monocular vision-based approach in the current study is described in this section. Figure \ref{fig:Flowchart}, shows the integration of aforementioned modules. The SLAM Module sends the scaled odometry to the Context-Awareness Module. The Context-Awareness Module processes the images using a scene segmentation scheme and generates a 1-D array. This array indicates the obstacle boundary which are inputs to the Mapping Module. The Control Module receives the control commands from a joystick and passes them to the Raspberry Pi. Robot Operating System (ROS) \cite{ROS} is used in this research to simplify the data exchange process between multiple modules.

\begin{figure*}[bt]
\centering
\includegraphics[width=.8\textwidth]{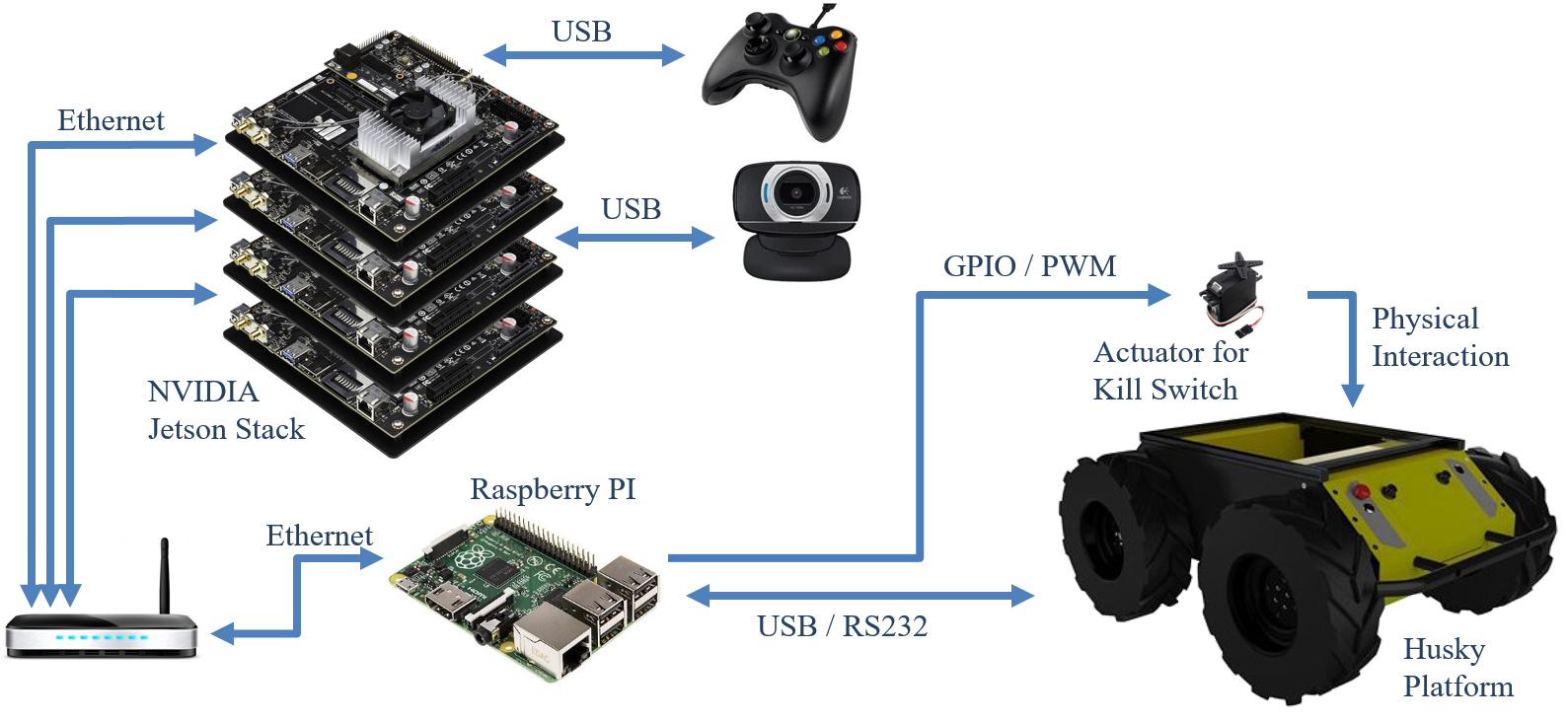}
\caption{Physical diagram of components in the platform. The channels used for interactions between the different physical modules are labeled in blue.}
\label{fig:Hardware}
\end{figure*}

Figure \ref{fig:Hardware}, illustrates the hardware used in the proposed system which are; A Clearpath Husky A200 \cite{Clearpath}, NVIDIA Jetson TX1 \cite{NVIDIA}, Raspberry Pi, Wi-fi enabled router, and a Microsoft Xbox controller. The controller is connected to the Control Module and is responsible for manually controlling the robot for initial mapping of the scene. This happens by sending this information to the Raspberry Pi through a TCP socket and the Raspberry Pi sends back the wheel encoder information. The Raspberry Pi can also operate a kill switch to stop the motors in case of an emergency. In Figure \ref{fig:Hardware}, There are four NVIDIA Jetson TX1 boards. Each module runs on one board to minimize logistics and integration time. A network via router on the robot connects all the Jetson boards. 

\section{System Description}
The main goal of this research is the integration between SLAM Monocular Module, Context-Awareness Module, Mapping Module, and Control Module. This integration leads to the generation of an occupancy grid map of the environment which forms a spatial map of obstacles and traversable space of the scene. In this section, other capabilities of these modules are explained in detail.
\subsection{Real-time visual SLAM with scaled odometry}
\begin{figure}[!htbp]
\centering
\includegraphics[width=0.35\textwidth]{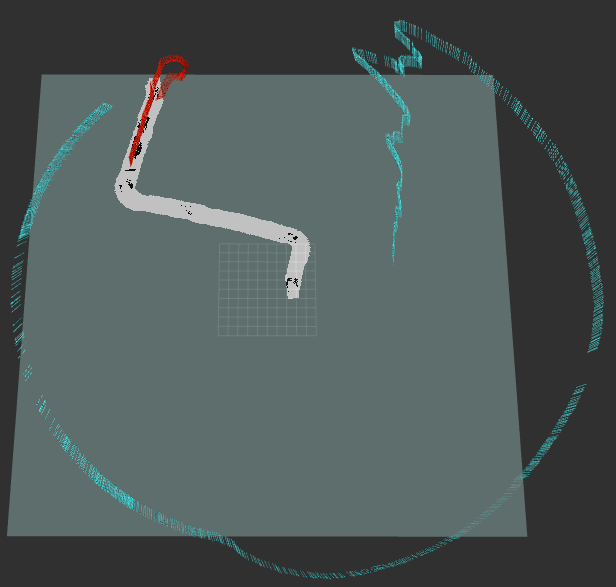}
\caption{Scaled ORB SLAM odometry (red) vs. unscaled ORB SLAM odometry (blue)}
\label{fig:SLAM}
\end{figure}

The provided odometry to the Context-Awareness Module cannot be used directly. To solve this issue, the proposed approach in  \cite{horn1987closed} is implemented to get the scaled information between visual odometry and wheel odometry. Both odometry are calculated for the entire path and then a closed-form solution is generated. Finally, The scaling matrix between wheel odometry and visual odometry is found and updates every second. The final odometry output that other modules can use is published as a ROS topic at a frequency of 1 Hz, as shown in Figure \ref{fig:overview}. Figure \ref{fig:SLAM}, shows scaled ORB-SLAM in red compared to unmodified SLAM in blue which shows that the scaled ORB-SLAM improves the trajectory. The unmodified SLAM is not able to detect the simple turn or even moving in the straight line.
\subsection{SLAM and Context-Awareness Modules contribution to provide segmented images}
The next step is to provide images and synchronize scaled odometry to the Context-Awareness Module for the segmentation of ground plane and obstacles. Contextual Awareness Module provides the intelligence and aids in decision-making while in motion. It strives to improve a general awareness of the environment by enhancing the visual information from the monocular camera.

Creating the segmented images and preparing them to be used by Mapping Module consists of three steps; pixel-wise semantic segmentation, filtering the segmentation vector, and perspective transformation. The segmentation model Enet \cite{paszke2016enet}, is used to produce a pixel-wise semantic segmentation map per image. The segmentation vector is a 1-D vector along the horizontal axis that represents the distance to the closest object at each point. Finally, a perspective transformation is implemented to convert from image to the world coordinate system. This information is used by Mapping Module to create the occupancy grid map of the environment.

\subsubsection{Pixel wise semantic segmentation}
The pixel-wise labeling task assigns a label to every pixel of the image. This typically requires models that are computationally heavy, with a lot of parameters. Due to the limited computing capability of the Jetson, a smaller model (ENet) is chosen with a slightly lower accuracy, but fast enough for near real-time performance. ENet method for semantic segmentation \cite{paszke2016enet} is designed to work on embedded boards and in the current research this method is implemented on the Jetson TX1 with image input size of $512\times 256$, at speed of 10 fps (much faster than other models such as Segnet and FCN \cite{badrinarayanan2017segnet,long2015fully}). 

\begin{figure}[H]
\centering
\includegraphics[width=0.45\textwidth]{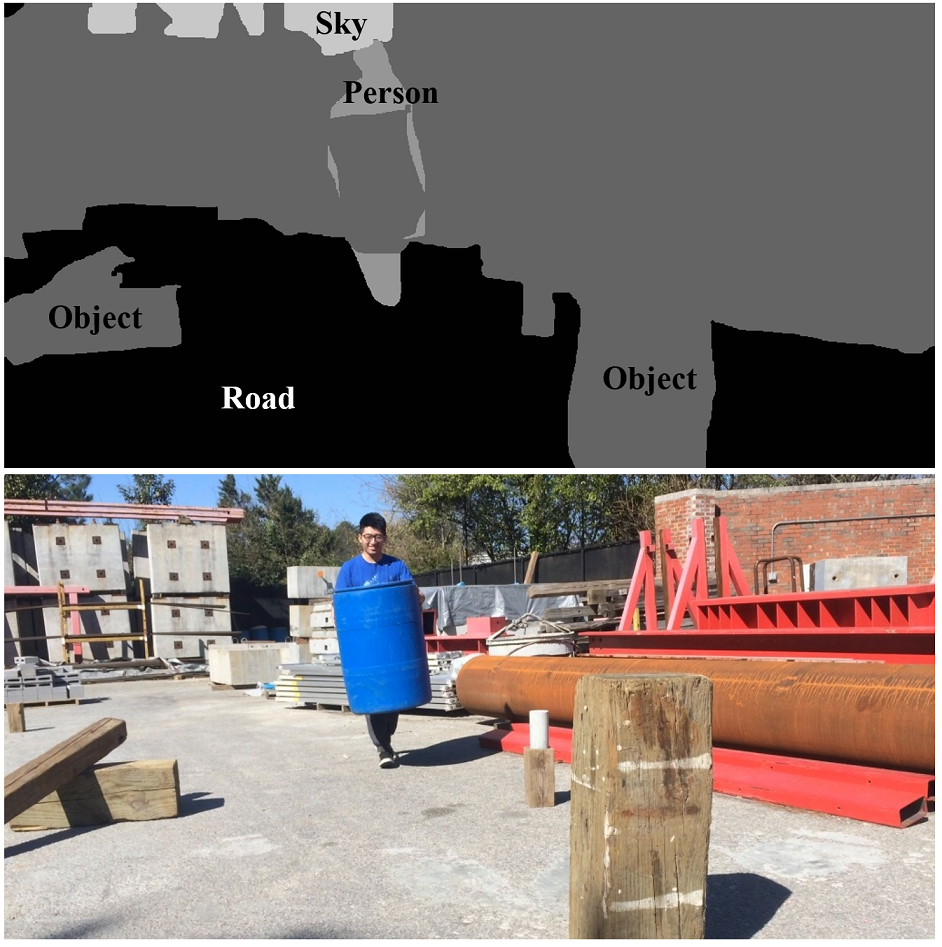}
\caption{Different labeled classes}
\label{fig:classes}
\end{figure}

The labeled pixel-wise data are input to ENet during the training phase. In this research, a scripting file is implemented to easily label images
with freehand drawing. 1000 images, captured from multiple
videos, are sampled and labeled manually. Grey-scale images with pixel information being the class label and size similar to input image size are used for training the network. Five labels are used for labeling as shown in Figure \ref{fig:classes}: object, road, person, sky, and unlabeled. 

The training process includes 2 steps. First, training the encoder part. The input images for encoder training has a size of $512\times 256$ and the size of output labeled map is $64\times 32$. The model is trained
for 300 epochs with a batch size of 10. Second, training the decoder part on top of the encoder to convert the intermediate map into the same dimensions as the full image. 
\subsubsection{Filtering the segmentation vector}
\begin{figure}[H]
\centering
\includegraphics[width=0.5\textwidth]{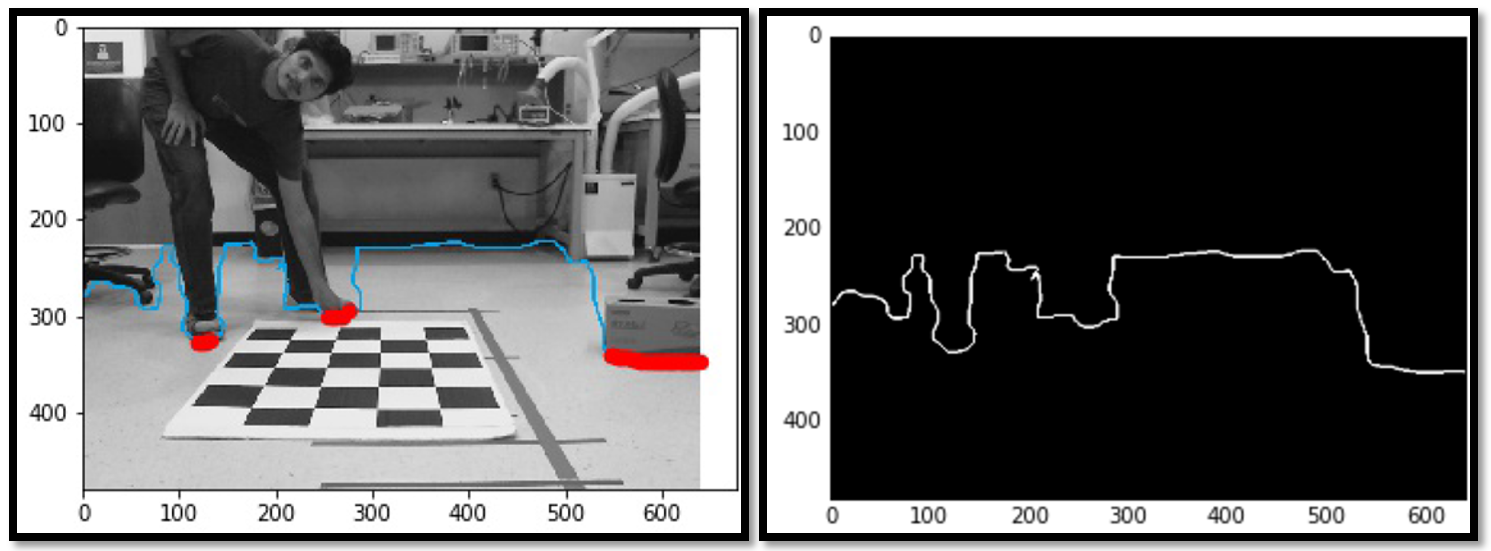}
\caption{Red dots indicated filtered points to be considered obstacles}
\label{fig:filter}
\end{figure}

The robot can move safe in the places that are known as road in the image. First of all, a 1-D vector is computed which provides the first instance of the obstacle when going through the bottom (the blue line in Figure \ref{fig:filter}). The filter is tuned to only include objects that are within 2.5 meters of the camera. Also when the vector is vertical it probably means that it is not actually the base of the object so it should not be included in the occupancy grid. To this end, points with the high gradient (more than one) in the x-direction are filtered out. The red segments in Figure \ref{fig:filter}, indicate filtered segments which are plotted as obstacles.  

\subsubsection{Perspective transform}
\begin{figure}[H]
\centering
\includegraphics[width=0.50\textwidth]{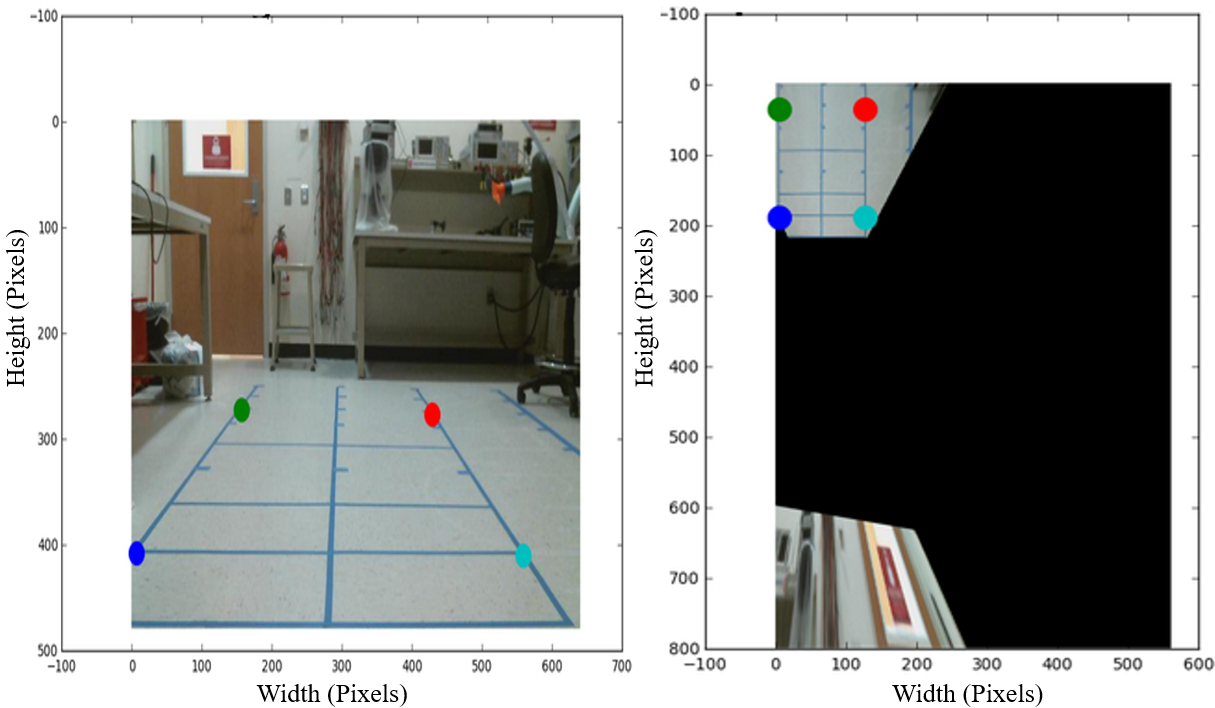}
\caption{Left: Original image, Right: Perspective transform}
\label{fig:transformation}
\end{figure}

The 2D image coordinates are transferred into physical distance values from the base of the robot using perspective transform. In the proposed system, the location and orientation of the camera are fixed related to the robot. Hence, a fixed perspective transform from specific focal length and camera position can be used for the whole process of mapping the pixel locations to its real-world locations (see Figure \ref{fig:transformation}). 

In reality, the perspective transform matrix is little different with the transformation matrix in Figure \ref{fig:transformation}. The reason is, the origin starts from the top of the image, not the camera itself. Also, it is necessary for the pixel mapping to be symmetric on the left and the right sides of the robot, but the camera is not exactly in the center of the rectangle.

\subsection{Occupancy grid map creation}
\begin{figure}[!htbp]
\centering
\includegraphics[width=0.35\textwidth]{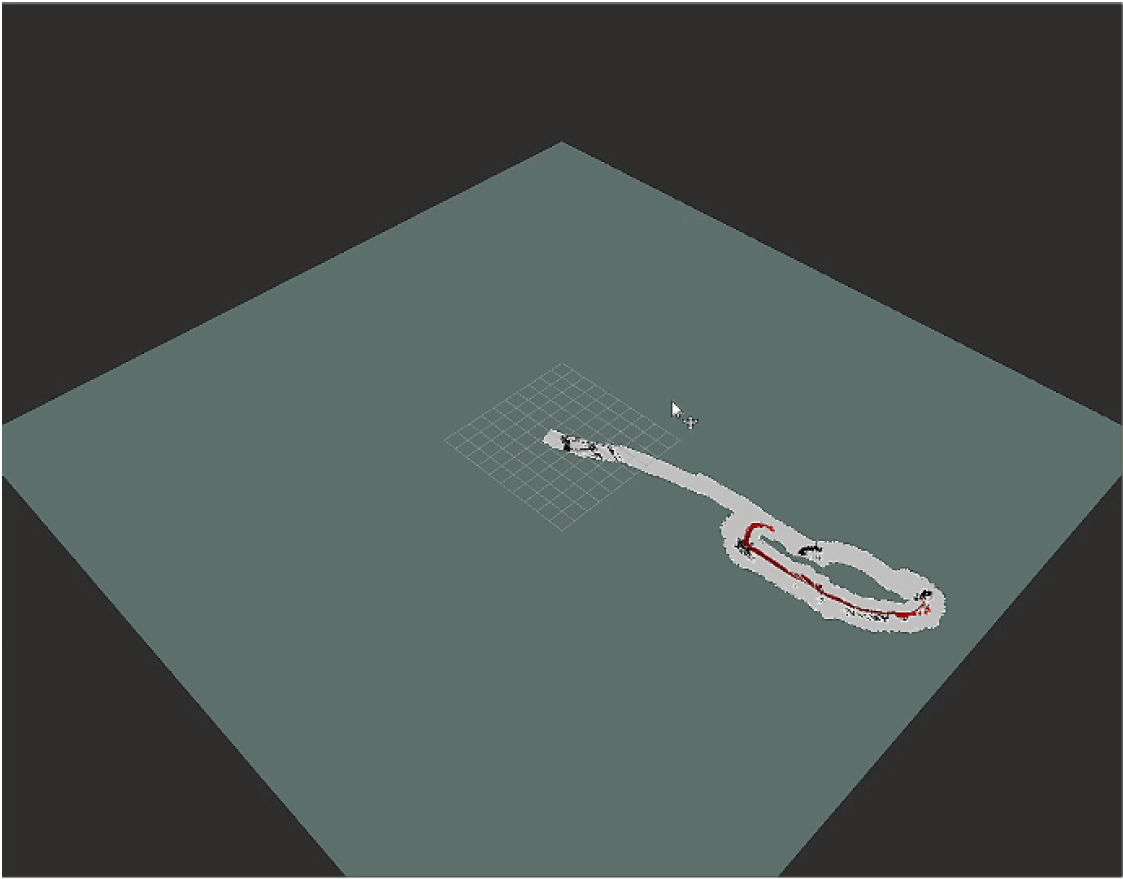}
\caption{Global map}
\label{fig:global}
\end{figure}

The Mapping Module uses the segmentation results to provide an occupancy grid map. The aforementioned obstacle position vector from Context-Awareness Module is processed to find the lower boundary of close obstacles. Next, the perspective transformation matrix helps to find the real world obstacle locations. Finally, the position of obstacles is plotted on a local rectangular map with 1.1m wide and 2.5m long and the global map incrementally updates by using the local map (see Figure \ref{fig:global}).

\subsection{Publishing tracking state of ORB-SLAM for SLAM and Control Modules integration}

It is necessary for the SLAM Module to know whether or not ORB-SLAM has initialized tracking. The Context-Awareness Module is not able to segment the images if the SLAM Module loses track. Also, the Mapping Module needs the images from SLAM Module to update the global map and the SLAM Module can only send the images in the tracking state. SLAM Module publishes following states: waiting for images, not initialized, tracking, and tracking lost. Providing this information for Control Module enables the robot to retrace its path in case tracking was lost. \cite{ORB_SLAM_Tracking_State} shows the tracking state in the ORB-SLAM Module.
\section{System Evaluation and Results}
\begin{figure*}[ht]
\centering
\includegraphics[width=\textwidth]{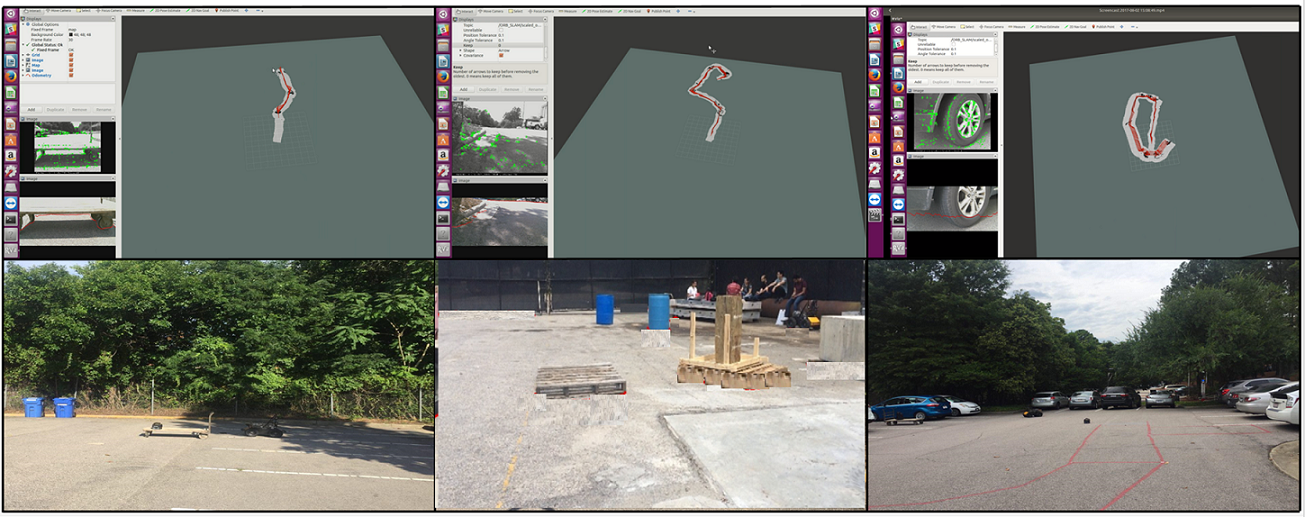}
\caption{Image of the environment (bottom) and its corresponding occupancy map (top).}
\label{fig:environment}
\end{figure*}

The proposed system is tested in three different outdoor environments with various object and weather conditions as shown in Table \ref{table:description}. 
The Figure \ref{fig:environment} shows a representative image of the environment and a screenshot of the occupancy grid map during its generation. The RVIZ tool of ROS is used to visualize the occupancy grid map. The red line in the image is the trajectory of the robot and the grey rectangles represent the mapped areas. 
\begin{table*}[ht]
\centering
\caption{Environment description}
\label{table:description}
\begin{tabular}{ccccc}
\hline
\begin{tabular}[c]{@{}c@{}}Evironment Type\end{tabular} & \begin{tabular}[c]{@{}c@{}}Object Type\end{tabular}                        & \begin{tabular}[c]{@{}c@{}}Weather \\ Condition\end{tabular} & \begin{tabular}[c]{@{}c@{}}Video Length\\  (minutes)\end{tabular} & \begin{tabular}[c]{@{}c@{}}Number of processed \\ frames (pipeline rate of 1 Hz)\end{tabular} \\ \hline
Parking space 1                                                & Car, curb                                                                         & Cloudy                                                       & 12.7                                                              & 762                                                                                              \\ 
Construction site                                              & \begin{tabular}[c]{@{}c@{}}Wooden planks,\\ trash bin, cement slab\end{tabular} & Cloudy                                                       & 15.15                                                             & 909                                                                                              \\ 
Parking scene 2                                                & \begin{tabular}[c]{@{}c@{}}Trash bins, utility cart\end{tabular}                & Sunny                                                        & 9.5                                                               & 570                                                                                              \\ \hline
\end{tabular}
\end{table*}

The focus of the validation in this research is on the integrated robotic system which brings multiple components together and runs in real-time. The videos of the whole pipeline demonstrate the capabilities of this integrated system in near real-time \cite{First,Second,Third}.

\begin{figure}[H]
\centering
\includegraphics[width=0.35\textwidth]{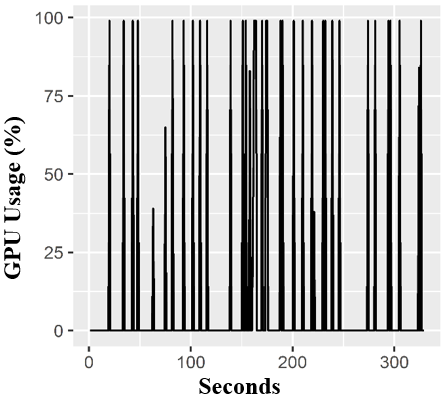}
\caption{Context-Awareness Module GPU usage graph (\%)}
\label{fig:gpu}
\end{figure}

The computational load on the Jetson boards is also presented for future systems and pipelines (see Table \ref{table:HardwareUsage}). Since the Jetson board has a quad-core processor, the percentages for the CPU are within a range of zero to 400. The CPU usage of ORB-SLAM is very high, but the scaling process in SLAM Module requires light computing. Figure \ref{fig:gpu}, shows that the Context-Awareness Module runs heavily on the GPU.  When an image is passed over to the ENet network, a segmented image is produced and is shown as a spike in the GPU usage in Figure \ref{fig:gpu}. This process publishes the boundary-position vector at the end of processing. The hardware usage for Mapping Module is not significant and it shows the potential of implementing more than one module on one Jetson. 
\begin{table}[H]
\caption{CPU usage statistics. (100 corresponds to full usage of one core)}
\label{table:HardwareUsage}
\centering
\renewcommand{\arraystretch}{2}

\begin{tabular}{l l}
\hline
\multicolumn{1}{c}{Code/CPU Usage} &
\multicolumn{1}{c}{Average CPU Usage} \\
\hline
ORB SLAM & 186 \\
odometry scaler & 5 \\
segmentation process & 101 \\
store\_images & 33 \\
global Map & 54 \\
local Map & 18  \\
\hline
\end{tabular}
\normalsize
\end{table}
      
\section{Conclusion}
This paper presents an integrated mobile robotic system that runs multiple vision-based components in real-time. The proposed system implements monocular SLAM and contextual understanding of a scene, which creates a 2D spatial map with detected obstacles. This system showcases the importance of a modular framework which can include latest SLAM and Context-Awareness algorithms in a plug and play format. This system is effective and can be run in real-time on multiple embedded platforms that are integrated as a system. The proposed system is a step forward in making intelligent and contextually aware robots ubiquitous. The results also demonstrate the potential for enabling a computer vision system for autonomous navigation. 

Some of the possible extensions and improvements to this projects are documented as follows. For instance, the proposed system does not have an effective way to deal with a large area to be mapped in real-time. A potential solution is to remove older parts of the global map and will be investigated in the authors' future work. 

The use of higher resolution input images with more features will result in better tracking and mapping. To ensure that the algorithm still runs in real-time, the feature extraction and matching parts can be moved to the GPU. This will allow us to get better results in a dynamic environment.

The computational load put on each Jetson board shows that the module corresponding to the segmentation task runs heavily on the GPU. The size of the model and high memory usage along with the need for real-time performance restricts the speed of the Husky \cite{Realtime_segmentation}. Improving the segmentation model in order to reduce the computational load in the Context Awareness Module will address this issue.

\bibliographystyle{ieeetr}
\bibliography{ISARC}
\end{document}